\documentclass[11pt]{article}

\usepackage[spanish,english]{babel}
\usepackage{arxiv}
\usepackage{algpseudocode}
\usepackage{algorithm}
\usepackage{graphicx}
\usepackage[english]{babel}
\usepackage{amsmath}
\usepackage{longtable}
\usepackage{siunitx}
\usepackage{booktabs}
\usepackage{subcaption}
\usepackage{adjustbox}
\usepackage{etoolbox} 
\usepackage[utf8]{inputenc} 
\usepackage[T1]{fontenc}    
\usepackage{url}            
\usepackage{booktabs}       
\usepackage{amsfonts}       
\usepackage{nicefrac}       
\usepackage{microtype}      
\usepackage{lipsum}
\usepackage{color}
\usepackage{amssymb}
\usepackage{float}
\usepackage{multirow}
\usepackage{afterpage}

\usepackage{booktabs} 
\usepackage{multirow} 
\usepackage{makecell} 
\usepackage{geometry} 
\usepackage{lineno}
\usepackage{xspace}
\usepackage{doi}
\usepackage{datetime}

\title{KairosHope: A Next-Generation Time-Series Foundation Model for Specialized Classification via Dual-Memory Architecture}

\author{
Luis Balderas\\
  {\small Department of Computer Science}\\
  {\small and Artificial Intelligence}\\
  {\small DiCITS, iMUDS, DaSCI}\\
  {\small University of Granada, Granada, Spain 18071}\\
  {\small Advanced Medical Imaging Group,}\\
  {\small Instituto de Investigación Biosanitaria de Granada (ibs.Granada)}\\
  \texttt{luisbalru@ugr.es} \\
\And
  José Alberto Rodríguez\\
    {\small Department of Computer Science}\\
  {\small and Artificial Intelligence}\\
  {\small DiCITS, iMUDS, DaSCI}\\
  {\small University of Granada, Granada, Spain 18071}\\
  \texttt{josealberto99@correo.ugr.es} \\
\And
  Miguel Lastra \\
  {\small Department of Software Engineering}\\
  {\small DiCITS, iMUDS, DaSCI}\\
  {\small University of Granada, Granada, Spain 18071}\\
  \texttt{mlastral@ugr.es} \\ 
\And
  Antonio Arauzo-Azofra \\
  {\small Department of Rural Engineering,}\\
  {\small DiCITS, iMUDS,}\\
  {\small University of Córdoba, Córdoba, Spain 14005}\\
  \texttt{arauzo@uco.es} \\ 
\And
  José M. Benítez\\
  {\small Department of Computer Science}\\
  {\small and Artificial Intelligence}\\
  {\small DiCITS, iMUDS, DaSCI}\\
  {\small University of Granada, Granada, Spain 18071}\\
  {\small Advanced Medical Imaging Group,}\\
  {\small Instituto de Investigación Biosanitaria de Granada (ibs.Granada)}\\
  \texttt{J.M.Benitez@decsai.ugr.es} \\
}

\begin{document}

\maketitle

\begin{abstract}
Time Series Foundation Models (TSFMs) have demonstrated notable success in general-purpose forecasting tasks; however, their adaptation to specialized classification problems remains constrained by the computational bottleneck of standard attention and the systematic omission of classical statistical knowledge. This technical report introduces KairosHope, a next-generation TSFM designed to reconcile massive generalization with analytical precision in classification tasks. The core of the proposal is the HOPE Block, an architecture that replaces quadratic attention with a dual-memory system: Titans modules for dynamic short-term retention and a Continuum Memory System (CMS) for the abstraction of long-term historical context.

To enrich the inductive bias, a Hybrid Decision Head is introduced, which fuses deep latent representations with deterministic statistical features extracted via \textit{tsfeatures}  \cite{tsfeatures}. KairosHope undergoes self-supervised pre-training on the massive Monash archive, combining Masked Time Series Modeling (MTSM) and contrastive learning (InfoNCE). Its subsequent adaptation to the UCR benchmark datasets is conducted through a rigorous Linear Probing and Full Fine-Tuning (LP-FT) protocol to prevent catastrophic forgetting. Empirical results demonstrate superior performance in domains characterized by strict temporal causality—such as HAR or Sensor data. Consequently, KairosHope establishes a robust and efficient framework for the adaptation of foundation models to time series analysis.
\end{abstract}

\section{Introduction}

Time series analysis and classification represent a central challenge across multiple disciplines, ranging from medical vital sign monitoring \cite{khan2025tracking} to anomaly detection in the Internet of Things (IoT) \cite{8926446} and financial analysis \cite{10.1145/3729531}. Unlike natural language processing (NLP) or computer vision, real-world time series exhibit a highly heterogeneous nature, characterized by variable-length sequences, non-stationary distributions, and complex interactions between local dynamics (short-term volatility) and global dependencies (macro-seasonality).

Historically, time series classification was dominated by approaches based on elastic distances, such as Dynamic Time Warping (DTW) \cite{berndt1994using}, as well as dictionary and shapelet-based algorithms \cite{lines2012shapelet}. The advent of deep learning radically transformed the field. Architectures such as Fully Convolutional Networks (FCN) \cite{wang2017time} and, most prominently, InceptionTime \cite{ismail2020inceptiontime}, established new performance standards by demonstrating a superior capacity to extract invariant local patterns without the need for manual feature engineering. However, these traditional deep learning models operate under an isolated paradigm: they must be initialized and trained from scratch for each specific dataset, rendering them incapable of transferring structural knowledge across domains.

Recently, the success of large language models (LLMs) has catalyzed the development of Time Series Foundation Models (TSFMs). Seminal proposals such as PatchTST \cite{nie2022time} demonstrated the efficacy of segmenting series into patches to preserve local semantics and reduce computational complexity. Concurrently, industry models such as TimesFM \cite{das2023decoder} and MOIRAI \cite{woo2024unified} scaled pre-training on massive data collections to achieve generalized forecasting capabilities. Throughout the last year, research has vigorously expanded toward general latent understanding and multi-domain classification, highlighting architectures such as UDE \cite{wang2025time}, which integrate Koopman operators and universal delay projections, or proposals like UniShape \cite{liu2026unified}, which attempt to incorporate geometric interpretability (shapelets) into massive architectures.

Despite this rapid evolution, the adaptation of the current generation of TSFMs to specialized classification tasks continues to present critical limitations. The standard attention mechanism suffers from a quadratic memory bottleneck that limits deep historical context, and the vast majority of purely neural architectures completely discard the valuable statistical knowledge inherent in classical time series theory.

To address this gap, this paper introduces KairosHope, a next-generation Time Series Foundation Model designed specifically to reconcile massive generalization with analytical precision in specialized classification. KairosHope provides three fundamental contributions. First, the HOPE Block \cite{behrouz2025nestedlearningillusiondeep} replaces traditional quadratic attention with a dual-memory architecture capable of modeling long-term dependencies with linear computational cost. Second, a dual-objective self-supervised pre-training regime based on the vast and heterogeneous Monash repository \cite{godahewa2021monash} forces the model to learn robust and invariant representations. Finally, a Hybrid Decision Head, combined with the LP-FT protocol, fuses deep embeddings with deterministically extracted statistical features, ensuring efficient adaptation to classification domains evaluated on the UCR benchmark \cite{8894743}. 

Throughout this document, the underlying architecture, the self-supervised training strategy, and the empirical results on the UCR benchmark datasets are detailed, demonstrating how KairosHope establishes a new paradigm in the adaptation of foundation models for time series classification.

\section{Our proposal}

This section introduces the proposed foundation model for time series classification: KairosHope. The architecture of KairosHope is designed to address the inherent challenges of TSFM, namely variable sequence lengths, non-stationarity, and the necessity of modeling both local semantics and long-term historical dependencies. The following subsections detail both the architecture and the training strategy employed.

\subsection{Architecture}

KairosHope processes sequences of diverse natures, origins, and types through a meticulously designed architecture comprising a preprocessing module, an HOPE based encoder \cite{behrouz2025nestedlearningillusiondeep}, and a classification head utilizing a hybrid strategy. Figure \label{fig:system} depicts the KairosHope architecture

\begin{figure}
  \includegraphics[width=\linewidth]{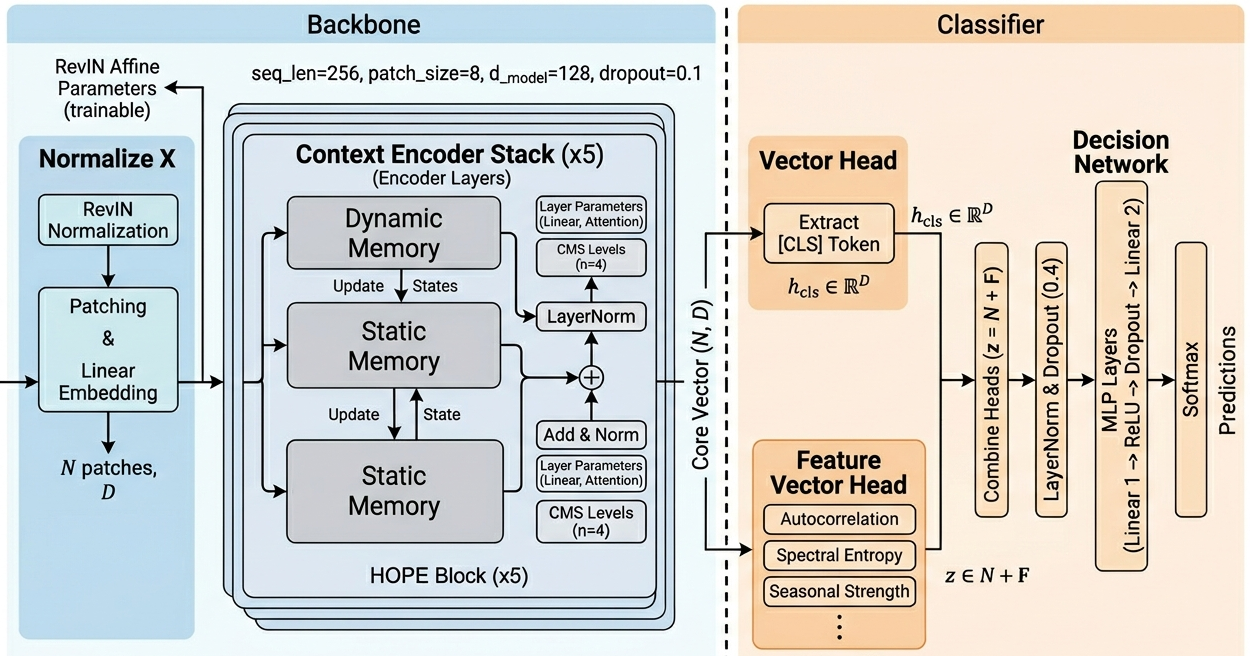}
  \caption{KairosHope architecture}
  \label{fig:system}
\end{figure}

\subsubsection{Preprocessing: Data Serialization, Patching, and RevIN}

Time series data encountered in academic and industrial contexts often exhibit significant variability and distributional shifts. To mitigate these issues and achieve normalization, KairosHope first applies Reversible Instance Normalization (RevIN) \cite{kim2021reversible} to the original input sequence $X \in \mathbb{R}^{L\times C}$, where $L$ denotes the sequence length and $C$ represents the number of channels. RevIN normalizes the sequence to zero mean and unit variance using trainable affine parameters, effectively removing non-stationary macro-trends temporarily.

Once normalized, the sequence is divided into non-overlapping segments, or patches, of length $P$. This patching mechanism maps the sequence point-wise into a sequence of $N$ tokens, where $N = \lfloor L/P \rfloor$. The utilization of patches offers two critical benefits:

\begin{itemize}
    \item Semantic extraction: It captures local temporal semantics, analogous to words in natural language sentences.
    \item Computation efficiency: It reduces the computational complexity of subsequent sequence modeling from $\mathcal{O}(L^2)$ to $\mathcal{O}(N^2)$.
\end{itemize}

Finally, these patches are linearly projected into an embedding space of dimension $D$, concatenated with a special [CLS] token, and augmented with positional encodings.

\subsection{HOPE Encoder}

Traditional Transformers \cite{wolf-etal-2020-transformers} rely on static-window self-attention \cite{vaswani2023attentionneed}, a mechanism that struggles when processing datasets with infinite horizons, primarily due to quadratic memory bottlenecks. To address these limitations, the core of the model replaces standard attention with the HOPE Block, a dual-memory system that integrates two advanced neural memory primitives:

\begin{itemize}
    \item Self-Modifying Titans: This module acts as the model's short to medium-term working memory. By processing patches in sequential chunks, the Titans \cite{behrouz2024titanslearningmemorizetest} module dynamically updates its internal memory state. This enables the network to adapt to sudden volatility and local temporal dynamics without the need to maintain an exhaustive attention matrix over the entire sequence.
    \item Proper Continuum Memory System (CMS): Operating in conjunction with the Titans module, the CMS is responsible for long-term abstraction. It compresses and persists historical representations across hierarchical levels. This mechanism ensures that macro-seasonal trends and distant historical context are preserved, thereby granting KairosHope a substantially extended receptive field.
\end{itemize}

Together, Titans and CMS modules map the input patch embeddings into deep, contextualized representations $H \in \mathbb{R}^{(N+1)\times D}$.

\subsubsection{Hybrid Classification Head}

Although deep representations are extremely powerful, completely disregarding certain traditional statistical features often leads to suboptimal performance in highly structured time series. KairosHope introduces a Hybrid Decision Head that integrates deep neural embeddings with deterministic time series features.

First, the latent representation is extracted from the encoder. For classification tasks, this corresponds to the state of the [CLS] token, $h_{\text{cls}} \in \mathbb{R}^{D}$. In parallel, the original time series is processed by a specialized feature extractor, \textit{tsfeatures} \cite{tsfeatures}, which computes a vector of statistical priors $f_{\text{ts}} \in \mathbb{R}^{K}$ that encompasses properties such as autocorrelation, spectral entropy, and seasonal strength, among others.

The final representation $z$ is the concatenation of both the deep embedding and the statistical features:

$$z = [~h_{\text{cls}}~ \|  ~f_{\text{ts}}~]$$

This vector $z$ is subsequently passed through a normalization layer, dropout, and finally a linear layer to produce the final classification labels.

\subsection{Training Strategy}

The training of KairosHope is governed by a two-stage paradigm designed to transform a general-purpose architecture into a high-precision predictor for specific domains. This progression enables the model to first learn the  ``universal laws'' of time series in a self-supervised setting, and subsequently achieve efficient specialization.

\subsubsection{Pre-training KairosHope}

Pre-training is conducted on a massive, heterogeneous corpus using a dual learning objective. This phase does not require external labels and focuses on the structural and relational understanding of the data.

\begin{enumerate}
    \item \textbf{Masked Time Series Modeling (MTSM).} Drawing on advancements in natural language processing and inspired by \cite{BERT}, a masked patch modeling strategy is implemented. During training, 40\% of input patches are randomly masked. The objective is to reconstruct the normalized values of these missing patches. To manage corpus heterogeneity, a dual-masking strategy is employed: a stochastic training mask for learning and a padding mask, which ensures the model ignores the artificial zeros introduced to align sequences of varying lengths. The reconstruction loss is defined by the Mean Squared Error (MSE):

    $$\mathcal{L}_{\text{MTSM}} = \frac{1}{M} \sum_{i \in M} ||\hat{P}_i - P_i||^2$$

    where $M$ represents the set of masked patches, $\hat{P}_i$ denotes the reconstructed patch, and $P_i$ represents the original patch.

    \item \textbf{Contrastive Learning.} To enhance the robustness of global representations, an InfoNCE \cite{rusak2025infonceidentifyinggaptheory} contrastive loss is integrated. Two augmented views of the same time series are generated by injecting Gaussian noise (2\%) and applying random scaling. This technique compels the model to distinguish between different augmented versions of the same time series (positive pairs) and other distinct series within the same batch (negative pairs). By maximizing latent similarity between representations of the same instance using a temperature of $\tau=0.1$, the model learns invariant features that are robust to perturbations and fundamental for downstream tasks. The loss is formulated as:

    $$\mathcal{L}_{\text{NCE}} = - \sum_{i} \log \frac{\exp(\text{sim}(z_i, z_j)/\tau}{\sum_k\exp(\text{sim}(z_i, z_j)/\tau}$$

\end{enumerate}

The total pre-training loss function is a weighted combination of both objectives.
$$\mathcal{L}_{\text{total}} = \lambda_1~\mathcal{L}_{\text{MTSM}} + \lambda_2 ~ \mathcal{L}_{\text{NCE}}$$

\subsubsection{Fine-tuning KairosHope}

Once the HOPE encoder has developed a universal understanding of temporal dynamics, fine-tuning is conducted on the target dataset utilizing the Linear Probing - Full Fine-tuning (LP-FT) strategy \cite{kumar2022finetuning}. This approach is vital for ensuring the retention of generalized representations throughout the fine-tuning process.

\begin{enumerate}
    \item \textbf{Linear Probing (LP).} Initially, the encoder weights are frozen, and only the hybrid decision head is trained. During this phase, the model learns to map rich, pre-trained representations to the specific classes or values of the target domain. Maintaining the encoder intact prevents initial gradient noise from degrading the memory structures learned within the Titans and CMS modules.
    \item \textbf{Full Fine-tuning (FT).} Upon initial convergence of the decision head, all model parameters are unlocked for comprehensive training. During this stage, a significantly lower learning rate is employed. This facilitates fine-grained adjustments to the encoder weights, specializing the system's memory toward the nuances and particularities of the target dataset while preserving the underlying generalist knowledge base.
\end{enumerate}

This dual LP-FT strategy ensures that KairosHope inherits the stability of the pre-training phase while achieving state-of-the-art precision in the specific target task.

\section{Empirical analysis}

To validate the efficacy of KairosHope and its generalization capability from a self-supervised environment to specific classification tasks, a thorough experimental framework has been designed. This section details the architectural configuration, optimization hyperparameters, datasets utilized, and the evaluation scheme for the results.

\subsection{Metrics}

To rigorously assess the performance of KairosHope across its different training phases, both unsupervised and supervised evaluation metrics are employed. During the foundational self-supervised pre-training phase, the model is evaluated using two primary loss-derived metrics:

\begin{itemize}
    \item Masked Time Series Modeling Mean Squared Error ($\text{MSE}_{\text{MTSM}}$): Quantifies the reconstruction accuracy of the masked temporal patches, measuring the backbone's ability to capture underlying deterministic dynamics and statistical regularities.
    \item InfoNCE Contrastive Loss ($\mathcal{L}_{\text{NCE}}$): Evaluates the quality of the learned latent space by measuring the model's proficiency in maximizing the similarity of augmented views of the same instance while segregating negative pairs within the batch.
\end{itemize}

With respect to the classification metrics, accuracy and macro-averaged F1-Score have been used.

\subsection{Network hyperparameters}

The base model architecture was configured with specific structural hyperparameters to balance expressive capacity with computational efficiency. The input sequence length is set to 256 time steps, while the patch size is established at 8 time steps, yielding a sequence of 32 patches per instance. The latent space dimension is fixed at 128, and the encoder depth comprises five HOPE blocks. Regarding the memory architecture, the CMS utilizes four hierarchical levels to facilitate long-term context compression. Finally, to ensure robust regularization, a dropout probability of 0.1 is applied within the backbone HOPE blocks to preserve the retention of underlying features, whereas a dropout rate of 0.4 is employed in the hybrid classifier head to effectively mitigate overfitting.

\subsection{Training hyperparameters}
The optimization process utilizes AdamW as the primary optimizer, leveraging its efficient decoupling of weight decay. Training follows the progression outlined in Section 3, with distinct configurations applied to each phase. During the pre-training phase, the model is trained for 50 epochs with a learning rate and weight decay of $1\times10^{-4}$, employing a batch size of 128.

For the fine-tuning stage, governed by the LP-FT protocol, a CrossEntropyLoss function mitigated by label smoothing with a factor of 0.1 is used to penalize overconfidence in model predictions. Phase 1, involving linear probing, is conducted over 15 epochs with a learning rate and weight decay of $1\times10^{-3}$. Subsequently, Phase 2, which consists of full fine-tuning, is performed over 30 epochs and employs differentiated, layer-wise learning rates of $1\times10^{-5}$ for the backbone and $1\times10^{-4}$ for the classifier. To ensure convergence, a CosineAnnealingLR scheduler gradually decays the learning rate to a lower bound of $1\times10^{-6}$. Finally, a batch size of 16 is adopted for this stage.

\subsection{Datasets}

To ensure that the model assimilates a wide variety of temporal dynamics, two distinct data collections are employed for the foundation and specialization phases. The pre-training corpus consists of massive, forecasting-oriented datasets extracted from the Monash Time Series Forecasting Repository \cite{godahewa2021monash}, specifically \textit{m4\_hourly, weather, electricity, traffic}, and \textit{tourism\_monthly}. This selection exposes the model to patterns ranging from high-frequency and extreme volatility to low-frequency seasonal macro-trends. For the fine-tuning corpus, the BakeOff classification benchmark \cite{Ruiz2021}, based on datasets from the UCR Time Series Classification Archive \cite{8894743, UCRArchive2018}, served as the reference for downstream evaluation. To align this evaluation with typical and pragmatic scenarios in both industrial and academic contexts, the experiment focused on a filtered subset of problems satisfying two conditions: possessing fewer than eight classes and containing sequences with fewer than 400 time steps per instance.

\subsection{Results}

The following section presents the experimental results for KairosHope. As shown in the appendix, Table \ref{results}, the proposed model was evaluated using the datasets containing fewer than eight classes and fewer than 400 time steps per instance from the Bake Off benchmark datasets, achieving a mean test accuracy of 80.11\% and an F1-score of 0.786. Table \ref{type}, which presents the mean accuracy categorized by dataset type using the aforementioned filter, indicates that HAR datasets yielded the highest performance (91.333\%), followed by SENSOR (89.6\%) and SPECTRO (83.67\%) datasets. Conversely, IMAGE datasets exhibited the lowest performance (66.21\%).

Besides, the empirical evaluation of KairosHope across varying dataset characteristics --namely the number of classes, time series length, and training set size-- reveals distinct performance trends, as summarized in Tables \ref{class}, \ref{length}, and \ref{ts}. When examining performance based on problem complexity via the number of target classes ($C$) in Table \ref{class}, KairosHope performs optimally on binary classification tasks ($C = 2$), achieving a Mean Accuracy of 83.91\% and a Mean F1-Score of 0.83. As the problem complexity scales to multi-class scenarios ($C > 2$), metrics experience a noticeable drop to a Mean Accuracy of 76.03\% and an F1-Score of 0.73. This degradation is a standard phenomenon in classification models, reflecting the heightened difficulty of distinguishing between a larger number of decision boundaries.

In terms of structural sequence traits, the model displays a non-linear relationship with time series length ($L$), as shown in Table \ref{length} The highest predictive capability is observed on medium-long sequences ($136 < L \le 235$) with a Mean Accuracy of 82.72\% and a Mean F1-Score of 0.81, closely followed by short sequences ($L \le 80$) at 80.32\% accuracy. However, performance dips slightly for ultra-long sequences ($L > 235$) to 77.19\%, indicating that extremely extended timelines might introduce noisy features or pose challenges for capturing long-range temporal dependencies. Conversely, the evaluation by training set size ($TS$) in Table \ref{ts} demonstrates a more linear, predictable trajectory. Aside from a minor performance fluctuation in the mid-range bracket ($30 < TS \le 70$), KairosHope scales highly effectively with data volume. Performance peaks decisively on the largest datasets ($TS > 400$), securing a Mean Accuracy of 83.77\% and a Mean F1-Score of 0.82. (Note: The final two rows of the first column in Table 4 contain a minor typographical error, using the variable $L$ instead of $TS$.) Taken together, these findings indicate that while KairosHope is a robust framework across diverse constraints, it delivers its most powerful performance when deployed on binary classification problems, provided with ample training samples, and restricted to moderately long time horizons.

\begin{table}[]
\centering
\begin{tabular}{l S}
\hline
{\textbf{Type}} & {\textbf{Mean Accuracy}}                              \\ \hline
{DEVICE}        & { 68.55}                                            \\ \hline
{ ECG}           & { 79.25} \\ \hline
{ HAR}           & { 91.33}                                            \\ \hline
{ IMAGE}         & { 66.21}                                            \\ \hline
{ MOTION}        & { 81.61}                                            \\ \hline
{ SENSOR}        & { 89.60} \\ \hline
{ SIMULATED}     & { 79.47}                                            \\ \hline
{ SPECTRO}       & { 83.67} \\ \hline
\end{tabular}
\caption{KairosHope results by dataset type}
\label{type}
\end{table}

\begin{table}[]
\centering
\begin{tabular}{l S S}
\hline
{\textbf{Number of Classes (C}} & {\textbf{Mean Accuracy}}   & {\textbf{Mean F1-Score}}                           \\ \hline
{$C=2$}        & { 83.91} & {0.83}                                            \\ \hline
{ $C>2$}           & { 76.03} & {0.73} \\ \hline

\end{tabular}
\caption{KairosHope results dividing datasets depending on the number of classes}
\label{class}
\end{table}

\begin{table}[]
\centering
\begin{tabular}{l S S}
\hline
{\textbf{Time Series Length (L)}} & {\textbf{Mean Accuracy}}   & {\textbf{Mean F1-Score}}                           \\ \hline
{$L \leq 80$}        & { 80.32} & {0.78}                                            \\ \hline
{ $80<L\leq136$}           & { 75.31} & {0.73} \\ \hline
{ $136<L\leq 235$}           & { 82.72} & {0.81} \\ \hline
{ $L> 235$}           & { 77.19} & {0.76} \\ \hline

\end{tabular}
\caption{KairosHope results dividing datasets depending on the time series length}
\label{length}
\end{table}

\begin{table}[]
\centering
\begin{tabular}{l S S}
\hline
{\textbf{Train Size (TS)}} & {\textbf{Mean Accuracy}}   & {\textbf{Mean F1-Score}}                           \\ \hline
{$TS \leq 30$}        & { 78.47} & {0.77}                                            \\ \hline
{ $30<TS\leq 70$}           & { 75.32} & {0.74} \\ \hline
{ $70<TS\leq 400$}           & { 80.16} & {0.8} \\ \hline
{ $TS> 400$}           & { 83.77} & {0.82} \\ \hline

\end{tabular}
\caption{KairosHope results dividing datasets depending on the train size}
\label{ts}
\end{table}

\section{Discussion}

The analysis of the empirical results obtained from the UCR benchmark following the pre-training and fine-tuning (LP-FT) phases provides critical insights into the inductive biases of the model. Generally, the integration of the Titans and CMS modules within the backbone demonstrates a notable superiority in Sensor, Motion, and ECG datasets, where temporal causality and long-term historical dependencies are the determining factors for successful classification. The hybrid decision head successfully capitalizes on the statistical priors of |\textit{tsfeatures} within these domains.

However, the analysis grouped by dataset typology reveals disparate behavior and a margin for improvement in the family of datasets categorized as IMAGE. In these problems, the one-dimensional time series do not represent a dynamic phenomenon over time; rather, they are the result of unwinding or extracting the spatial contour of two-dimensional objects from images.

To improve the generalization capability and performance of the model in this specific category, a data augmentation strategy was designed and implemented. Inspired by spatial shape processing literature, the training set was enriched by generating synthetic time series from images with morphological variations, following the contour extraction methodology described in \cite{CAO2000297}. The objective of this intervention was to provide the fine-tuning phase with a significantly larger volume of data to calibrate the network against rotations, translations, and scale variations intrinsic to spatial contours.

Despite this robust augmentation strategy, empirical results did not show a significant improvement in Accuracy or F1-Score metrics for IMAGE-type datasets, with performance remaining stagnant relative to the baseline without augmentation. This negative result is highly informative and suggests two primary hypotheses regarding the structural limitations of the approach in this particular niche:

\begin{itemize}
    \item  Incompatibility of Temporal Inductive Bias: The HOPE block, particularly through the CMS and the sequential processing of Titans, is heavily optimized to model strict temporal causality (where step $t$ depends on $t-1$). In $1D$ image contours, the starting and ending points are arbitrary and lack a real direction of causal flow, which confounds the long-term memory retention mechanisms designed for genuine time series.
    \item Misalignment of the Hybrid Head: The deterministic statistical feature extractor, which serves as a fundamental pillar in the classification head, computes metrics such as spectral entropy, trend strength, or seasonality. While these concepts possess deep semantic meaning in climatic or financial series, they lack representational utility when applied to the perimeter of a leaf or the silhouette of an insect.
\end{itemize}

Consequently, the performance discrepancy observed in IMAGE-type datasets demonstrates that conventional temporal priors are not directly extrapolatable to problems of geometric invariance, even when employing data augmentation techniques. This contrast underscores that the success of the architecture depends largely on the alignment between the model's inductive biases and the intrinsic nature of the underlying domain.

Finally, while we acknowledge that using the arithmetic mean of results across datasets with highly diverse characteristics is not entirely statistically rigorous, we believe it provides a suggestive indicator of the method's average behavior across a broad spectrum of use cases. The reader should keep this specific aggregation approach in mind to properly contextualize and evaluate the provided overall results.

\section{Conclusions}

In this technical report, KairosHope has been presented, a foundation model designed to transform time series classification through a dual-memory architecture and a two-phase training regime. The results obtained validate the central hypothesis that the integration of long-term (CMS) and short-term (Titans) memory mechanisms enables the capture of complex temporal dynamics that conventional models frequently overlook.

The primary conclusions derived from this development are summarized as follows:

\begin{itemize}
    \item Efficacy of the HOPE Architecture: The HOPE block has proven to be an efficient alternative to quadratic attention, facilitating linear processing that does not sacrifice the ability to model extensive historical dependencies.
    \item Value of the Hybrid Representation: The fusion of deep neural embeddings with deterministic statistical features (\textit{tsfeatures}) provides an essential anchor of robustness, which is particularly beneficial when fine-tuning data is limited.
    \item Robustness of the LP-FT Protocol: The Linear Probing strategy followed by Full Fine-Tuning is confirmed as a superior methodology for knowledge transfer from foundation models, achieving precise specialization without degrading the universal representations learned during pre-training.
    \item Limitations of Temporal Biases: Experimentation with IMAGE-type datasets and the utilization of synthetic data augmentation techniques reveal that the geometric nature of spatial contours requires inductive biases distinct from purely causal-temporal ones.
    \item Robust Scaling with Data Volume: KairosHope demonstrated excellent scalability, achieving its highest performance on large datasets ($\text{TS} > 400$) with a Mean Accuracy of 83.77\%. This indicates that the model effectively capitalizes on larger data volumes to learn robust representations.
    \item Sensitivity to Task Complexity: The framework proved highly effective for binary classification tasks (83.91\% Mean Accuracy) but saw a performance drop of approximately 7.8\% in multi-class scenarios ($\text{C} > 2$). Enhancing multi-class decision boundary separation remains a clear avenue for future optimization.
    \item Optimal Temporal Horizon: Evaluation across varying time series lengths revealed that KairosHope thrives on medium-long sequences ($136 < L \le 235$). However, the slight degradation observed in ultra-long sequences ($L > 235$) suggests a vulnerability to long-term noise or difficulties capturing extended temporal dependencies.
\end{itemize}

\section{Future Work}

Building upon these findings, future lines of research for the evolution of KairosHope will focus on the following areas:

\begin{enumerate}
    \item Geometric Invariance: Developing specific feature extraction modules for contour-type data that can be integrated into the hybrid head without compromising performance on dynamic series.
    \item Pre-training Scalability: Expanding the pre-training corpus toward even more massive and multimodal domains to observe the emergence point of zero-shot classification capabilities.
    \item CMS Optimization: Investigating more aggressive memory compression techniques to enable the processing of ultra-long sequences (comprising millions of time steps) on consumer-grade hardware.
    \item Multivariate Extension: Extending the KairosHope architecture to handle multivariate time series classification. This involves researching inter-channel correlation mechanisms and cross-dimensional attention to capture the complex dependencies between multiple variables without compromising the computational efficiency of the HOPE block.
\end{enumerate}

\section*{Acknowledgment}

This research has been partially supported by Proyecto PID2023-151336OB-I00 financiado por MICIU/AEI /10.13039/501100011033 y por FEDER, UE.

We do appreciate Prof. Keogh's comments on earlier versions of the paper, which have undoubtedly improved the content. Also, we would like to thank the creators and maintainers of the UCR Time Series Archive \cite{UCRArchive2018} and the Monash Time Series Forecasting Archive for making their comprehensive data repositories publicly available to the research community.

\section*{Appendix A: Complete Results}

\begin{longtable}{l S S}
    \toprule
    \textbf{Dataset} & {\textbf{Accuracy}} & {\textbf{F1-Score}} \\
    \midrule
    \endfirsthead
    
    \toprule
    \textbf{Dataset} & {\textbf{Accuracy}} & {\textbf{F1-Score}} \\
    \midrule
    \endhead
    
    \bottomrule
    \endfoot
    
ArrowHead&56.57142857142857&0.5641626100690162\\ \hline 
CBF&67.77777777777777&0.6640175026769933\\ \hline 
ChlorineConcentration&56.041667&0.443216 \\ \hline 
Coffee&96.428571&0.964331\\ \hline
DistalPhalanxTW&69.7841726618705&0.6110701063805654\\ \hline 
DistalPhalanxOutlineAgeGroup&73.38129496402878&0.7352169914934349\\ \hline 
DistalPhalanxOutlineCorrect&76.81159420289855&0.7590766643736147\\ \hline 
Earthquakes&75.53956834532374&0.6570710373849681\\ \hline 
ECG200&76.0&0.7493408134642356\\ \hline 
ECG5000&93.5111111111111&0.922910980797562\\ \hline 
ECGFiveDays&66.89895470383276&0.656779934691913\\ \hline 
ElectricDevices&68.55142004928025&0.6776415571890589\\ \hline 
FaceFour&57.95454545454545&0.5214180128814275\\ \hline 
GunPoint&91.33333333333333&0.912949938186097\\ \hline 
ItalyPowerDemand&93.58600583090379&0.9358347290674586\\ \hline 
Lightning7&67.12328767123287&0.6527953647212309\\ \hline 
MiddlePhalanxTW&58.44155844155844&0.5617732621519979\\ \hline 
MiddlePhalanxOutlineAgeGroup&60.38961038961039&0.567260440720805\\ \hline 
MiddlePhalanxOutlineCorrect&73.53951890034364&0.7157076352356141\\ \hline 
MoteStrain&87.1405750798722&0.8709911526051363\\ \hline 
NonInvasiveFetalECGThorax1&84.78371501272265&0.8458288551510422\\ \hline 
NonInvasiveFetalECGThorax2&87.43002544529261&0.8751505153518795\\ \hline 
PhalangesOutlinesCorrect&78.9044289044289&0.7816893825704399\\ \hline 
Plane&100.0&1.0\\ \hline 
ProximalPhalanxTW&75.60975609756098&0.719639225894391\\ \hline 
ProximalPhalanxOutlineAgeGroup&84.39024390243902&0.8478692311127316\\ \hline 
ProximalPhalanxOutlineCorrect&81.44329896907216&0.7971331732806101\\ \hline 
SonyAIBORobotSurface1&85.52412645590682&0.8556840290908578\\ \hline 
SonyAIBORobotSurface2&85.41448058761804&0.8556763115196905\\ \hline 
Strawberry&91.62162162162163&0.9170060314581605\\ \hline 
SyntheticControl&95.33333333333333&0.953147634009059\\ \hline 
ToeSegmentation1&85.52631578947368&0.8551373216557303\\ \hline 
ToeSegmentation2&77.6923076923077&0.7934071365655516\\ \hline 
Trace&99.0&0.9900372960372961\\ \hline 
TwoLeadECG&80.59701492537313&0.8043232019273321\\ \hline 
TwoPatterns&98.725&0.987245132091711\\ \hline 
UWaveGestureLibraryX&76.46566164154103&0.7579990347713328\\ \hline 
UWaveGestureLibraryY&65.52205471803462&0.6452369732639096\\ \hline 
UWaveGestureLibraryZ&66.97375767727526&0.6556171697192146\\ \hline 
UWaveGestureLibraryAll&88.80513679508654&0.8872015780005506\\ \hline 
Wafer&99.01038286826736&0.9900338776863941\\ \hline 
Wine&62.96296296296296&0.5707472178060413\\ \hline 
Worms&68.83116883116882&0.6587416784895778\\ \hline 
WormsTwoClass&77.92207792207792&0.779599539249565\\ \hline 
Yoga&70.9&0.6865083159819906\\ \hline 
    \caption{KairosHope results for UCR and Bake Off benchmark datasets}
    \label{results}
\end{longtable}

\bibliographystyle{unsrt} 
\bibliography{citas}

@misc{behrouz2025nestedlearningillusiondeep,
      title={Nested Learning: The Illusion of Deep Learning Architectures}, 
      author={Ali Behrouz and Meisam Razaviyayn and Peilin Zhong and Vahab Mirrokni},
      year={2025},
      eprint={2512.24695},
      archivePrefix={arXiv},
      primaryClass={cs.LG},
      url={https://arxiv.org/abs/2512.24695}, 
}

@article{CAO2000297,
title = {Digital hand atlas and web-based bone age assessment: system design and implementation},
journal = {Computerized Medical Imaging and Graphics},
volume = {24},
number = {5},
pages = {297-307},
year = {2000},
issn = {0895-6111},
doi = {https://doi.org/10.1016/S0895-6111(00)00026-4},
url = {https://www.sciencedirect.com/science/article/pii/S0895611100000264},
author = {F. Cao and H.K. Huang and E. Pietka and V. Gilsanz}
}

@inproceedings{kim2021reversible,
  title     = {Reversible Instance Normalization for Accurate Time-Series Forecasting against Distribution Shift},
  author    = {Kim, Taesung and 
               Kim, Jinhee and 
               Tae, Yunwon and 
               Park, Cheonbok and 
               Choi, Jang-Ho and 
               Choo, Jaegul},
  booktitle = {International Conference on Learning Representations},
  year      = {2021},
  url       = {https://openreview.net/forum?id=cGDAkQo1C0p}
}

@misc{behrouz2024titanslearningmemorizetest,
      title={Titans: Learning to Memorize at Test Time}, 
      author={Ali Behrouz and Peilin Zhong and Vahab Mirrokni},
      year={2024},
      eprint={2501.00663},
      archivePrefix={arXiv},
      primaryClass={cs.LG},
      url={https://arxiv.org/abs/2501.00663}, 
}

@inproceedings{wolf-etal-2020-transformers,
    title = "Transformers: State-of-the-Art Natural Language Processing",
    author = "Thomas Wolf and Lysandre Debut and Victor Sanh and Julien Chaumond and Clement Delangue and Anthony Moi and Pierric Cistac and Tim Rault and Rémi Louf and Morgan Funtowicz and Joe Davison and Sam Shleifer and Patrick von Platen and Clara Ma and Yacine Jernite and Julien Plu and Canwen Xu and Teven Le Scao and Sylvain Gugger and Mariama Drame and Quentin Lhoest and Alexander M. Rush",
    booktitle = "Proceedings of the 2020 Conference on Empirical Methods in Natural Language Processing: System Demonstrations",
    month = oct,
    year = "2020",
    address = "Online",
    publisher = "Association for Computational Linguistics",
    url = "https://www.aclweb.org/anthology/2020.emnlp-demos.6",
    pages = "38--45"
}

@misc{vaswani2023attentionneed,
      title={Attention Is All You Need}, 
      author={Ashish Vaswani and Noam Shazeer and Niki Parmar and Jakob Uszkoreit and Llion Jones and Aidan N. Gomez and Lukasz Kaiser and Illia Polosukhin},
      year={2023},
      eprint={1706.03762},
      archivePrefix={arXiv},
      primaryClass={cs.CL},
      url={https://arxiv.org/abs/1706.03762}, 
}

@Manual{tsfeatures,
  title = {tsfeatures: Time Series Feature Extraction},
  author = {Rob Hyndman and Yanfei Kang and Pablo Montero-Manso and Mitchell O'Hara-Wild and Thiyanga Talagala and Earo Wang and Yangzhuoran Yang},
  year = {2026},
  url = {https://pkg.robjhyndman.com/tsfeatures/},
}

@misc{BERT,
      title={BERT: Pre-training of Deep Bidirectional Transformers for Language Understanding}, 
      author={Jacob Devlin and Ming-Wei Chang and Kenton Lee and Kristina Toutanova},
      year={2019},
      eprint={1810.04805},
      archivePrefix={arXiv},
      primaryClass={cs.CL},
      url={https://arxiv.org/abs/1810.04805}, 
}

@misc{rusak2025infonceidentifyinggaptheory,
      title={InfoNCE: Identifying the Gap Between Theory and Practice}, 
      author={Evgenia Rusak and Patrik Reizinger and Attila Juhos and Oliver Bringmann and Roland S. Zimmermann and Wieland Brendel},
      year={2025},
      eprint={2407.00143},
      archivePrefix={arXiv},
      primaryClass={cs.LG},
      url={https://arxiv.org/abs/2407.00143}, 
}

@inproceedings{
kumar2022finetuning,
title={Fine-Tuning can Distort Pretrained Features and Underperform Out-of-Distribution},
author={Ananya Kumar and Aditi Raghunathan and Robbie Matthew Jones and Tengyu Ma and Percy Liang},
booktitle={International Conference on Learning Representations},
year={2022},
url={https://openreview.net/forum?id=UYneFzXSJWh}
}

@Article{Ruiz2021,
author={Ruiz, Alejandro Pasos
and Flynn, Michael
and Large, James
and Middlehurst, Matthew
and Bagnall, Anthony},
title={The great multivariate time series classification bake off: a review and experimental evaluation of recent algorithmic advances},
journal={Data Mining and Knowledge Discovery},
year={2021},
month={Mar},
day={01},
volume={35},
number={2},
pages={401-449},
issn={1573-756X},
doi={10.1007/s10618-020-00727-3},
url={https://doi.org/10.1007/s10618-020-00727-3}
}

@ARTICLE{8894743,
  author={Dau, Hoang Anh and Bagnall, Anthony and Kamgar, Kaveh and Yeh, Chin-Chia Michael and Zhu, Yan and Gharghabi, Shaghayegh and Ratanamahatana, Chotirat Ann and Keogh, Eamonn},
  journal={IEEE/CAA Journal of Automatica Sinica}, 
  title={The UCR time series archive}, 
  year={2019},
  volume={6},
  number={6},
  pages={1293-1305},
  doi={10.1109/JAS.2019.1911747}}

@InProceedings{godahewa2021monash,
              author = "Godahewa, Rakshitha and Bergmeir, Christoph and Webb, Geoffrey I. and Hyndman, Rob J. and Montero-Manso, Pablo",
              title = "Monash Time Series Forecasting Archive",
              booktitle = "Neural Information Processing Systems Track on Datasets and Benchmarks",
              year = "2021"
            }

@misc{UCRArchive2018,
        title = {The UCR Time Series Classification Archive},
        author = {Dau, Hoang Anh and Keogh, Eamonn and Kamgar, Kaveh and Yeh, Chin-Chia Michael and Zhu, Yan 
                  and Gharghabi, Shaghayegh and Ratanamahatana, Chotirat Ann and Yanping and Hu, Bing 
                  and Begum, Nurjahan and Bagnall, Anthony and Mueen, Abdullah and Batista, Gustavo and Hexagon-ML},
        year = {2018},
        month = {October},
        note = {\url{https://www.cs.ucr.edu/~eamonn/time_series_data_2018/}}
    }

@ARTICLE{8926446,
  author={Cook, Andrew A. and Mısırlı, Göksel and Fan, Zhong},
  journal={IEEE Internet of Things Journal}, 
  title={Anomaly Detection for IoT Time-Series Data: A Survey}, 
  year={2020},
  volume={7},
  number={7},
  pages={6481-6494},
  doi={10.1109/JIOT.2019.2958185}}

@article{10.1145/3729531,
author = {Arsenault, Pierre-Daniel and Wang, Shengrui and Patenaude, Jean-Marc},
title = {A Survey of Explainable Artificial Intelligence (XAI) in Financial Time Series Forecasting},
year = {2025},
issue_date = {October 2025},
publisher = {Association for Computing Machinery},
address = {New York, NY, USA},
volume = {57},
number = {10},
issn = {0360-0300},
url = {https://doi.org/10.1145/3729531},
doi = {10.1145/3729531},
journal = {ACM Comput. Surv.},
month = may,
articleno = {265},
numpages = {37}
}

@article{khan2025tracking,
  title={Tracking vital signs of a patient using channel state information and machine learning for a smart healthcare system},
  author={Khan, Muhammad Imran and Jan, Mian Ahmad and Muhammad, Yar and Do, Dinh-Thuan and Rehman, Ateeq ur and Mavromoustakis, Constandinos X and Pallis, Evangelos},
  journal={Neural Computing and Applications},
  volume={37},
  number={28},
  pages={23065--23079},
  year={2025},
  publisher={Springer}
}

@inproceedings{berndt1994using,
  title={Using dynamic time warping to find patterns in time series},
  author={Berndt, Donald J and Clifford, James},
  booktitle={Proceedings of the 3rd international conference on knowledge discovery and data mining},
  pages={359--370},
  year={1994}
}

@inproceedings{lines2012shapelet,
  title={A shapelet transform for time series classification},
  author={Lines, Jason and Davis, Luke M and Hills, Jon and Bagnall, Anthony},
  booktitle={Proceedings of the 18th ACM SIGKDD international conference on Knowledge discovery and data mining},
  pages={289--297},
  year={2012}
}

@inproceedings{wang2017time,
  title={Time series classification from scratch with deep neural networks: A strong baseline},
  author={Wang, Zhiguang and Yan, Weizhong and Oates, Tim},
  booktitle={2017 International joint conference on neural networks (IJCNN)},
  pages={1578--1585},
  year={2017},
  organization={IEEE}
}

@article{ismail2020inceptiontime,
  title={Inceptiontime: Finding alexnet for time series classification},
  author={Ismail Fawaz, Hassan and Lucas, Benjamin and Forestier, Germain and Pelletier, Charlotte and Schmidt, Daniel F and Weber, Jonathan and Webb, Geoffrey I and Idoumghar, Lhassane and Muller, Pierre-Alain and Petitjean, Fran{\c{c}}ois},
  journal={Data mining and knowledge discovery},
  volume={34},
  number={6},
  pages={1936--1962},
  year={2020},
  publisher={Springer}
}

@article{nie2022time,
  title={A time series is worth 64 words: Long-term forecasting with transformers. arXiv 2022},
  author={Nie, Yuqi and Nguyen, Nam H and Sinthong, Phanwadee and Kalagnanam, Jayant},
  journal={arXiv preprint arXiv:2211.14730},
  year={2022}
}

@article{das2023decoder,
  title={A decoder-only foundation model for time-series forecasting},
  author={Das, Abhimanyu and Kong, Weihao and Sen, Rajat and Zhou, Yichen},
  journal={arXiv preprint arXiv:2310.10688},
  year={2023}
}

@inproceedings{woo2024unified,
  title={Unified training of universal time series forecasting transformers},
  author={Woo, Gerald and Liu, Chenghao and Kumar, Akshat and Xiong, Caiming and Savarese, Silvio and Sahoo, Doyen},
  booktitle={Forty-first International Conference on Machine Learning},
  year={2024}
}

@article{wang2025time,
  title={A Time-Series Foundation Model by Universal Delay Embedding},
  author={Wang, Zijian and Tao, Peng and Shi, Jifan and Bao, Rui and Liu, Rui and Chen, Luonan},
  journal={arXiv preprint arXiv:2509.12080},
  year={2025}
}

@article{liu2026unified,
  title={A Unified Shape-Aware Foundation Model for Time Series Classification},
  author={Liu, Zhen and Wang, Yucheng and Li, Boyuan and Zheng, Junhao and Eldele, Emadeldeen and Wu, Min and Ma, Qianli},
  journal={arXiv preprint arXiv:2601.06429},
  year={2026}
}

\end{document}